\documentclass[10pt,twocolumn,letterpaper]{article}

\usepackage[pagenumbers]{cvpr} 

\providecommand{\TODO}[1]{}

\definecolor{cvprblue}{rgb}{0.21,0.49,0.74}
\usepackage[pagebackref,breaklinks,colorlinks,allcolors=cvprblue]{hyperref}

\def\paperID{*****} 
\def\confName{CVPR}
\def\confYear{2026}

\title{YILDIZ-VPR: A Novel Dataset with Dense Coverage Under Diverse Environmental Conditions for Visual Place Recognition}

\author{Serdar Yıldız\\
Department of Computer Engineering\\
Yildiz Technical University, Istanbul, Türkiye\\
BILGEM, TUBITAK, Gebze, Kocaeli, Türkiye\\
{\tt\small serdar.yildiz@std.yildiz.edu.tr}
\and
Abbas Memiş\\
Department of Computer Engineering\\
Istanbul University, Istanbul, Türkiye\\
{\tt\small abbas.memis@istanbul.edu.tr}
\and
Songül Varlı\\
Department of Computer Engineering\\
Yildiz Technical University, Istanbul, Türkiye\\
{\tt\small svarli@yildiz.edu.tr}
}

\begin{document}
\maketitle
\begin{abstract}
Visual Place Recognition (VPR) aims to recognize the location of a query image by comparing it with a set of geo-referenced images. Although many datasets have been proposed for VPR, collecting dense and diverse visual data from pedestrian-level viewpoints is still an important need. In this paper, we introduce YILDIZ-VPR, a visual geo-localization dataset collected through repeated walking traversals on the Davutpasa campus of Yildiz Technical University. The dataset includes outdoor scenes captured at different times of day, seasons, and weather conditions. It contains a wide range of visual content, including historical buildings, modern structures, roads, green areas, and wooded regions. Each video was recorded with a GoPro 9 camera and synchronized with GPS sensor data to provide location labels for the extracted frames. In addition to GPS coordinates, the dataset also includes auxiliary sensor information such as gyroscope, speed, and temperature data. With its dense coverage and long-term visual variability, YILDIZ-VPR provides a useful resource for studying image-based and temporal visual place recognition under realistic outdoor conditions. The YILDIZ-VPR dataset is available at \hyperlink{https://www.github.com/...}{https://www.github.com/...}
\end{abstract}    
\section{Introduction}

\begin{figure*}[t]
\centering
\includegraphics[width=\textwidth]{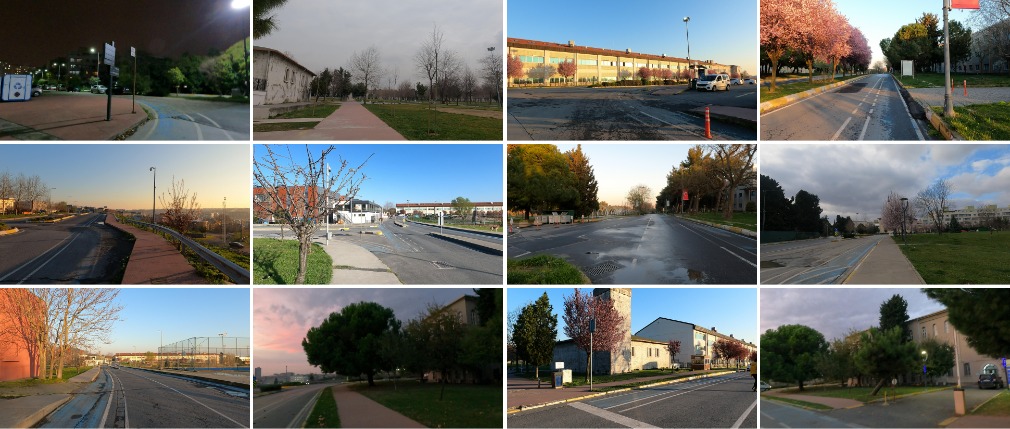}
\caption{Sample images from the YILDIZ-VPR dataset.}
 \label{Fig:dataset-samples}
 \end{figure*}

Visual Place Recognition (VPR) aims to identify the location of a visual observation by matching it against a set of previously collected geo-referenced images. It is commonly formulated as an image retrieval problem, where a query image is represented by a visual descriptor and compared with gallery images whose locations are known \cite{masone2021survey,zhang2021visual}. This formulation has made VPR an important component of visual localization systems, with applications in autonomous navigation, mobile robotics, augmented reality, and location-aware visual search.

Despite its practical importance, robust place recognition remains challenging in real-world environments. A single place may exhibit substantially different appearances depending on illumination, weather, season, time of day, viewpoint, shadows, and dynamic objects. These factors become particularly critical in dense localization scenarios, where visually similar locations may be only a few meters apart. Therefore, progress in VPR depends not only on stronger visual representations but also on datasets that reflect the spatial density and environmental variability observed in practical navigation settings.

GPS-based localization is widely used in modern positioning and navigation systems \cite{drawil2012gps}. However, GPS measurements may be noisy or unreliable in outdoor areas surrounded by buildings, trees, narrow paths, or other sources of signal degradation. In such cases, visual information can provide a complementary cue for localization. Rather than replacing GPS, VPR can support location estimation by using the visual structure of the environment, especially when accurate or stable coordinate measurements are difficult to obtain.

Several public datasets have contributed significantly to VPR research by providing large-scale, long-term, or cross-condition visual data. Early and widely used datasets such as St Lucia, Eynsham, San Francisco, Nordland, Tokyo 24/7, and Pitts250k introduced different forms of geographical coverage, viewpoint diversity, temporal variation, and illumination change \cite{StLucia,Eynsham,SanFrancisco,Nordland,Tokyo,Pitts250k}. Later datasets, including SPED, Oxford RobotCar, Mapillary SLS, SVOX, AmsterTime, GSV-CITIES, and San Francisco XL, further extended the field with long-term recordings, large-scale street-level imagery, historical views, multi-city coverage, and challenging environmental changes \cite{SPED,OxfRoboCar,Mapillary,svox,amstertime,ConvAP,CosPlace}. These datasets have shaped the evaluation of VPR methods and remain highly valuable for the community.

However, many existing datasets are collected using vehicle-mounted cameras, trains, surveillance cameras, or street-view platforms. While these acquisition strategies are effective for road-based and city-scale localization, they do not fully represent pedestrian-level visual perception. Road-centric data may provide limited access to walking paths, open spaces, building surroundings, green areas, and transitions between different outdoor regions. For applications involving pedestrians, wearable cameras, or mobile robots operating in human-scale environments, dense data collected from walking trajectories is therefore an important complementary resource.

Another challenge is the joint availability of dense spatial coverage and repeated observations under diverse environmental conditions. A dataset may cover a large geographical region, but this does not necessarily mean that the same locations are observed across different times of day, seasons, weather conditions, and viewpoints. Repeated observations of the same environment provide a more suitable basis for studying how visual place representations change over time. Such data are particularly useful for research on day-night matching, long-term localization, temporal VPR, and robustness under environmental variation.

To address these needs, we introduce \textbf{YILDIZ-VPR}, a dense visual place recognition dataset collected through repeated walking traversals. The dataset was recorded on the Davutpasa campus of Yildiz Technical University, which contains a diverse set of outdoor scenes, including historical buildings, modern structures, roads, green areas, open spaces, and wooded regions. This environment provides a compact yet visually rich setting, where natural and man-made structures coexist within a relatively dense spatial area.

\section{YILDIZ-VPR Dataset}
\label{sec:dataset}

YILDIZ-VPR was designed to provide dense, pedestrian-level visual data for visual place recognition research. The dataset focuses on outdoor place recognition under realistic environmental changes, rather than controlled or single-session image matching. For this purpose, repeated walking traversals were recorded in the same environment across different times of day, weather conditions, and seasonal periods. This design makes the dataset suitable for studying both spatially dense localization and long-term visual variation.

\begin{figure*}[t]
\centering
\includegraphics[width=\textwidth]{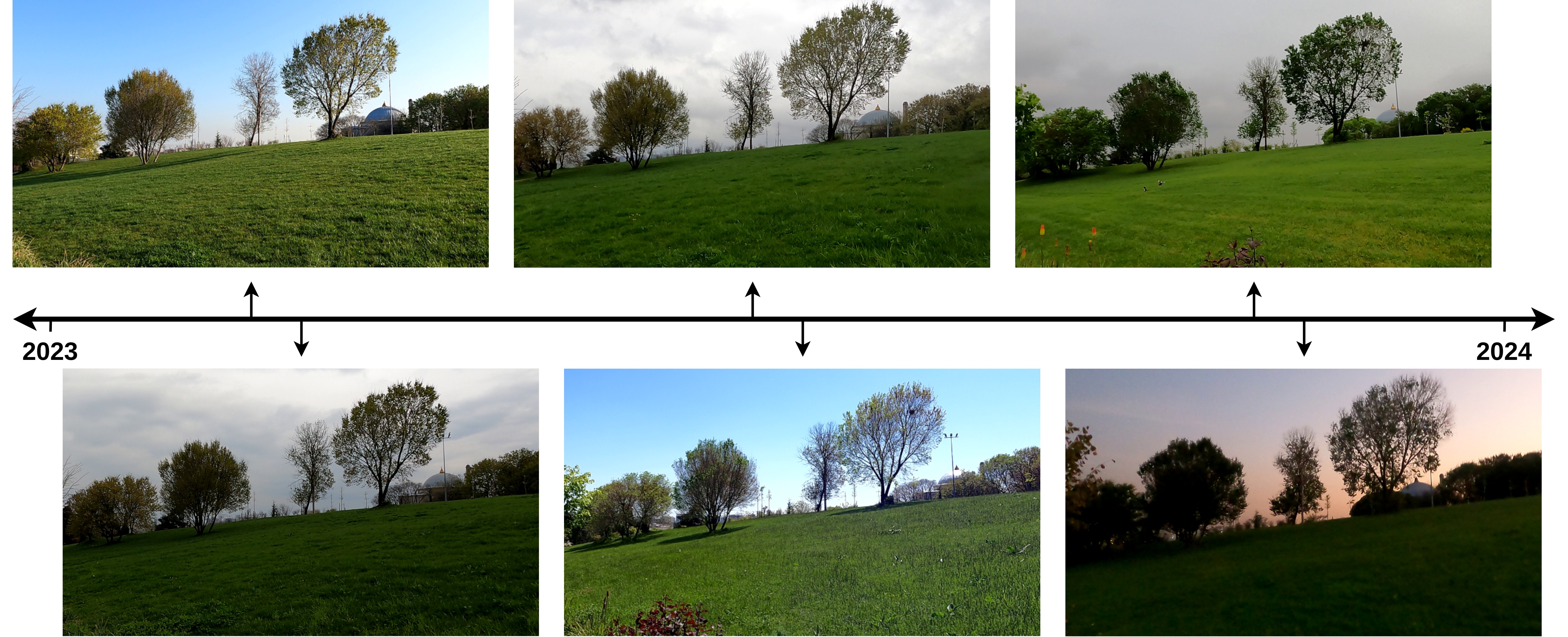}
\caption{Timeline and sample diversity of the YILDIZ-VPR dataset.}
\label{Fig:dataset-timeline}
\end{figure*}

\subsection{Collection Environment}

The dataset was collected on the Davutpasa campus of Yildiz Technical University. This location was selected because it contains a diverse mixture of outdoor scene types within a compact geographical area. The campus includes historical buildings, modern architectural structures, roads, open spaces, grass-covered areas, and wooded regions. These elements create a visually rich environment where natural and man-made structures appear together.

Indoor areas were not included in the dataset. This decision was made to keep the focus on outdoor visual place recognition, where environmental changes such as lighting, weather, shadows, and seasonal appearance differences play an important role. Since many outdoor locations on the campus are visually close to each other, the dataset also contains challenging cases in which nearby places may share similar visual patterns.

\subsection{Acquisition Protocol}

YILDIZ-VPR was recorded using a GoPro 9 camera during walking traversals. The videos were captured at 4K resolution and 30 frames per second. Unlike vehicle-mounted or street-view datasets, the camera viewpoint follows a pedestrian-level trajectory. This provides visual observations that are closer to how a person, wearable camera, or mobile robot perceives an outdoor environment.

The same areas were visited multiple times to capture the visual appearance of locations under different conditions. Some traversals were performed close to each other in time to increase viewpoint diversity, while others were recorded across different periods to capture changes in illumination, season, and weather. As a result, the dataset contains both fine-grained spatial coverage and repeated observations of the same environment under changing conditions.

\subsection{GPS-based Annotation}

Each recorded video is paired with GPS sensor data. The image frames were synchronized with the GPS measurements to assign geographical coordinates to the visual samples. Since GPS accuracy can be affected by buildings, trees, weather conditions, and other environmental factors, low-quality GPS measurements were filtered during dataset preparation. Only frames associated with GPS accuracy values below 5 meters were included in the dataset.

After filtering, the dataset provides location annotations with an average GPS accuracy of approximately 1.5 meters. The GPS signal was recorded approximately every 122 milliseconds, corresponding to a frequency of about 8--9 Hz. The videos were sampled in synchronization with the GPS stream at approximately 4 Hz. Depending on walking speed, this sampling strategy corresponds to roughly four images per meter, resulting in dense visual coverage of the traversed routes.

\subsection{Dataset Organization}

YILDIZ-VPR is organized into three main parts: training, daytime test, and nighttime test. This structure was designed to support the analysis of environmental and temporal changes in visual place recognition. The training set contains densely sampled frames from the main walking traversals. The daytime test set includes query images captured during daytime conditions, while the nighttime test set contains query images recorded at night.

\begin{figure*}[t]
\begin{subfigure}[t]{0.48\textwidth}
\centering
\includegraphics[width=\textwidth]{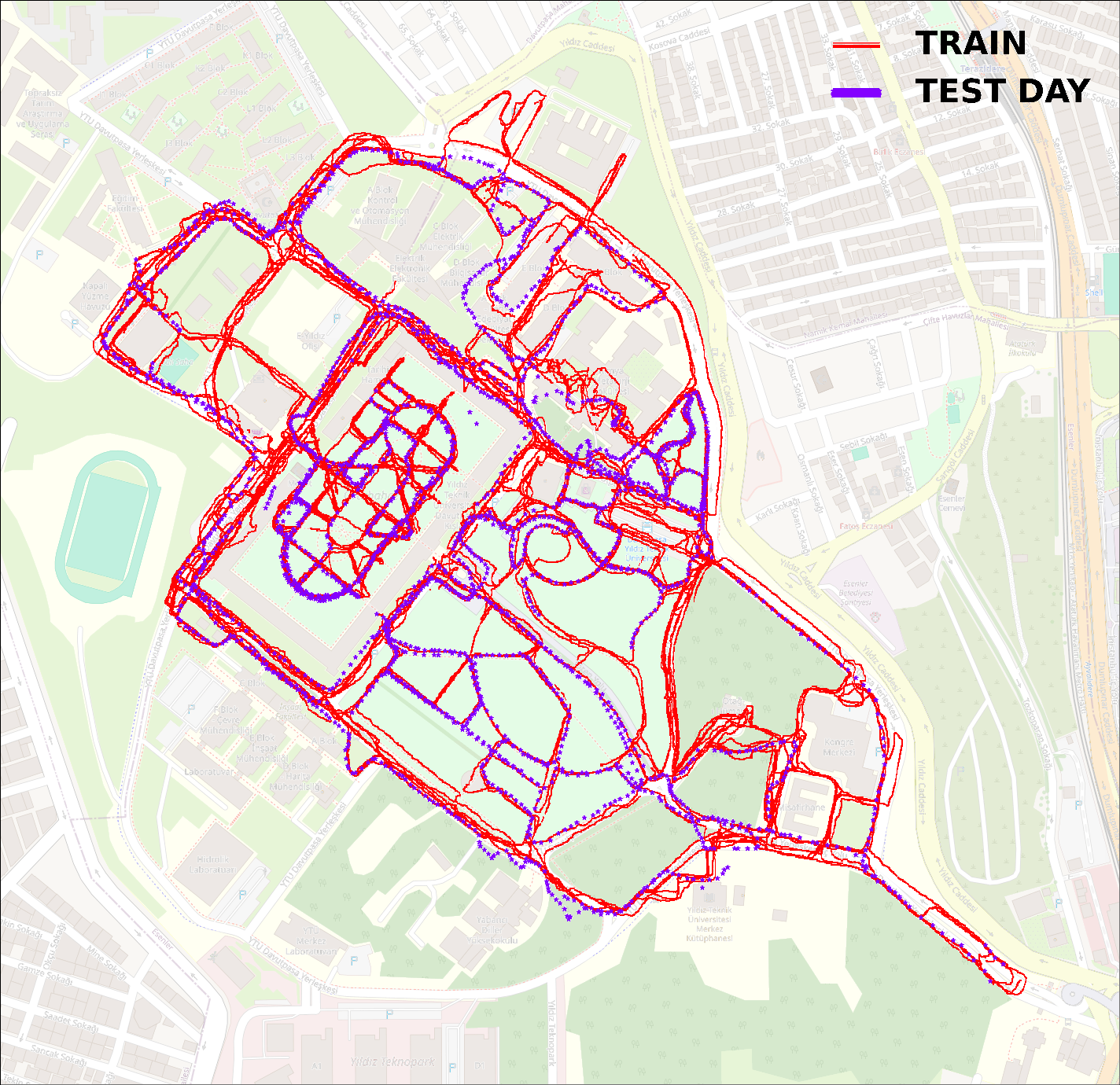}
\end{subfigure}
\hfill
\begin{subfigure}[t]{0.48\textwidth}
\centering
\includegraphics[width=\textwidth]{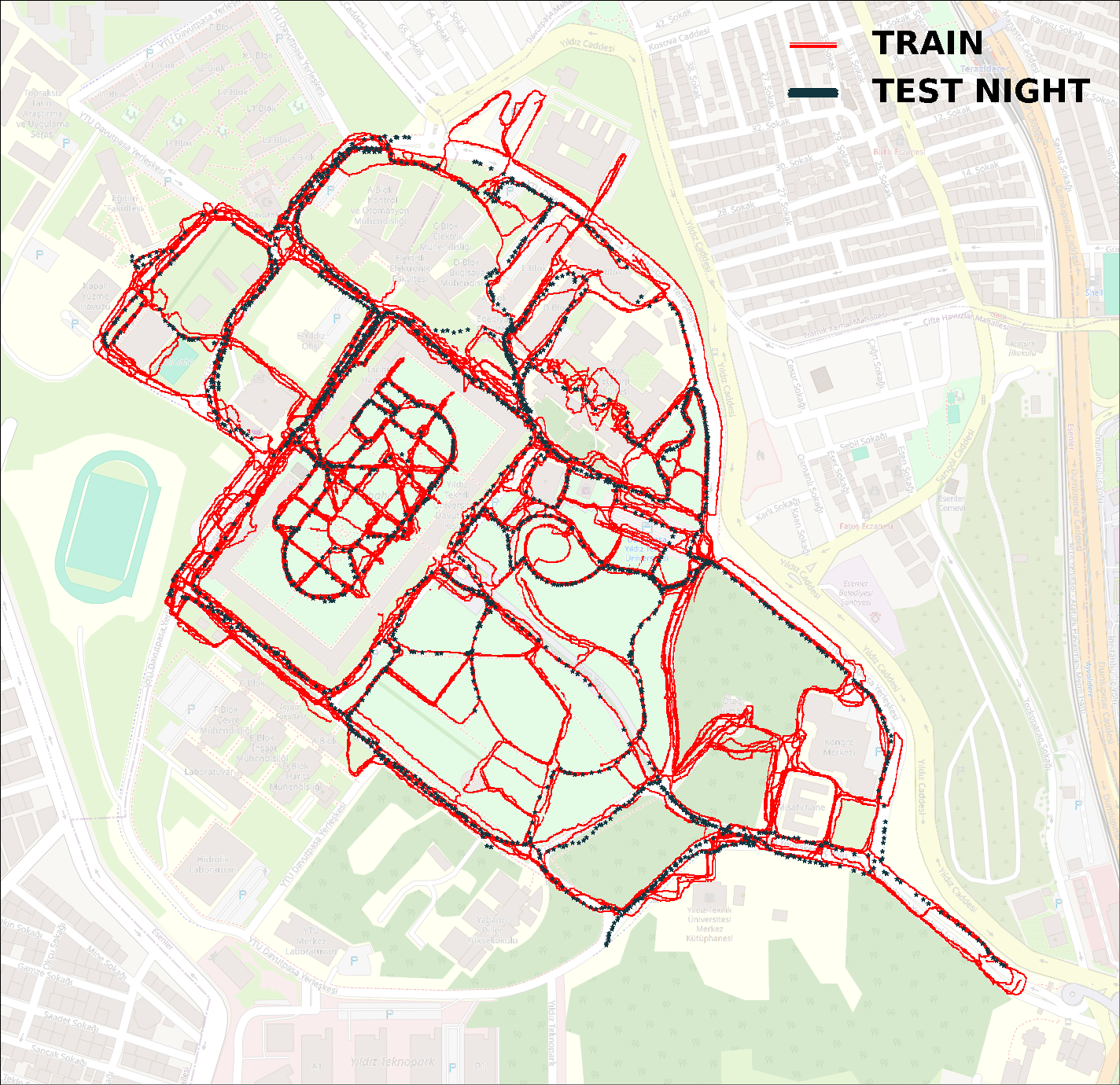}
\end{subfigure}
\caption{Spatial distribution of the YILDIZ-VPR dataset samples.}
\label{Fig:dataset-map}
\end{figure*}

The training frames were sampled from the recorded videos at regular intervals. In contrast, the query images in the test sets were manually selected to reduce redundancy between consecutive frames. This selection strategy increases the diversity of the query images and avoids constructing test sets from nearly identical adjacent frames.

The final dataset contains 370,464 training images, 2,377 daytime query images, and 1,902 nighttime query images. The training set can be used as the gallery set for image retrieval-based VPR experiments, while the daytime and nighttime query sets can be used to study localization under different illumination conditions.

\subsection{Metadata}

In addition to RGB image data and GPS coordinates, YILDIZ-VPR includes auxiliary sensor metadata. These metadata include sensor temperature, speed, and three-axis gyroscope measurements. Such information can provide additional context about the camera motion and acquisition conditions.

Although the dataset can be directly used for image-based VPR, the availability of video sequences and synchronized sensor data also makes it suitable for future temporal place recognition studies. For example, sequential models may use consecutive frames, motion cues, or sensor information to improve localization robustness under challenging visual conditions. Therefore, YILDIZ-VPR is not limited to single-image retrieval settings and can also support future research on temporal and sensor-aware visual localization.

\section{Conclusion}

In this work, we introduced YILDIZ-VPR, a densely sampled visual place recognition dataset collected through repeated pedestrian-level traversals of the Davutpasa campus of Yildiz Technical University under different times of day, seasons, weather conditions, and illumination settings. The dataset contains 370,464 training images, 2,377 daytime query images, and 1,902 nighttime query images, covering diverse outdoor scenes such as historical and modern buildings, roads, walking paths, open spaces, green areas, and wooded regions. Each image is associated with GPS coordinates. By combining dense spatial coverage, repeated observations of the same environment, pedestrian-level viewpoints, and substantial long-term appearance variation, YILDIZ-VPR offers a complementary resource for investigating image-based, sequential, day-to-night, and multimodal visual place recognition. Future work will establish comprehensive evaluation protocols and benchmark representative VPR methods on the dataset.

{
    \small
    \bibliographystyle{ieeenat_fullname}
    \bibliography{main}
}


\end{document}